\documentclass[letterpaper, 10 pt, conference]{ieeeconf}  

\IEEEoverridecommandlockouts                              

\usepackage{xcolor}
\usepackage{subcaption}   
\usepackage{tabularx}
\usepackage{multirow}
\usepackage{multicol}
\usepackage{ctable}
\usepackage{dcolumn}
\usepackage{balance}
\usepackage[T1]{fontenc}
\usepackage{graphicx}
\usepackage{color, soul, cite}
\usepackage[dvipsnames]{xcolor}

\makeatletter
\let\NAT@parse\undefined
\makeatother

\newcommand{\fref}[1]{Fig.~\ref{#1}}
\newcommand{\sref}[1]{Sec.~\ref{#1}}
\newcommand{\tref}[1]{Table~\ref{#1}}

\title{\LARGE \bf
When Should Robots Intervene? Balancing Engagement and Intrusiveness in Human-Robot Interaction

}

\author{Lavinia Hriscu$^{1,2}$, Valerio Bo$^{1,2}$, Alberto Sanfeliu$^{1,2}$ and Anais Garrell$^{1,2}$
  		\thanks{$^{1}$Institut de Robòtica i Informàtica Industrial (CSIC-UPC)
        {\tt\small \{name.surname\}@iri.upc.edu}}
        \thanks{$^{2}$ Universitat Politècnica de Catalunya (UPC), Barcelona, 08034, Spain.   }
\thanks{This work was supported by the European project TORNADO with grant number HORIZON-CL4-2024-DIGITAL-EMERGING-01-101189557 and
JST Moonshot R \& D grant number JPMJMS2011. The authors acknowledge Fernando Herrero Cotarelo for his technical support in this study.}
  \thanks{\copyright\ 2026 IEEE. Personal use of this material is permitted.
Permission from IEEE must be obtained for all other uses, in any current or future media, including reprinting/republishing this material for advertising or promotional purposes, creating new collective works, for resale or redistribution to servers or lists, or reuse of any copyrighted component of this work in other works.}
}

\usepackage{graphicx}
\usepackage{amsmath}

\begin{document}

\maketitle
\thispagestyle{empty}
\pagestyle{empty}

\begin{abstract}


Designing effective Human-Robot Interaction in task-oriented settings requires carefully balancing user engagement with socially acceptable levels of robot intrusiveness. In this paper, we examine how different robot intervention strategies shape user experience, interaction dynamics, perceived intrusiveness, and sense of support. We compare two approaches: a continuous engagement-seeking robot strategy, and a context-aware strategy that selectively intervenes based on the user’s state and task context. Both approaches rely on multimodal behavioral cues, including body orientation and attentional signals, to guide robot actions. We evaluate these strategies in a user study with 32 participants performing a task in a simulated hospital environment. Our findings show that higher interaction frequency does not necessarily lead to better engagement. Instead, we observe a systematic trade-off between perceived support and intrusiveness, influenced by factors such as physical proximity and user effort. These results provide empirical evidence that effective engagement in HRI depends on adaptive, context-sensitive intervention policies. 

\end{abstract}

\section{Introduction} \label{sec:introduction}

Improving the quality of Human-Robot Interaction (HRI) requires systems that incorporate users' needs into their decision-making processes. In many applications, robots can assist users in unfamiliar environments that require task-specific expertise. This is especially relevant in hospital and clinical settings, where end-users may rely on robotic systems to perform tasks. In such contexts, the effectiveness of interaction depends on the robot's ability to adapt to users who may be cognitively loaded or lacking domain expertise. 

One important indicator of interaction quality is engagement, defined in multiple ways in the literature~\cite{doherty2018engagement}. In this work, engagement is defined as a person’s willingness to continue interacting with a robot and is assessed through verbal and nonverbal cues, as well as contextual factors such as task, environment, and user expertise. 
In hospital settings, users often split attention between the robot and critical tasks, so reduced observable attention does not necessarily indicate reduced willingness to interact~\cite{corrigan2013social}.

Designing engagement methods is non-trivial: while interactive robots can enhance user excitement and perceived intelligence~\cite{schillaci2013evaluating}, prolonged interaction may lead to negative experiences~\cite{sorrentino2024investigating}. 
We treat engagement as a dynamic signal, one that robots should not aim to maximize continuously, but rather regulate it by timing their interventions appropriately and avoiding behaviors that may feel intrusive~\cite{ziadeh2023feeling}.

\begin{figure}[t]
    \centering
    \includegraphics[width=0.75\columnwidth]{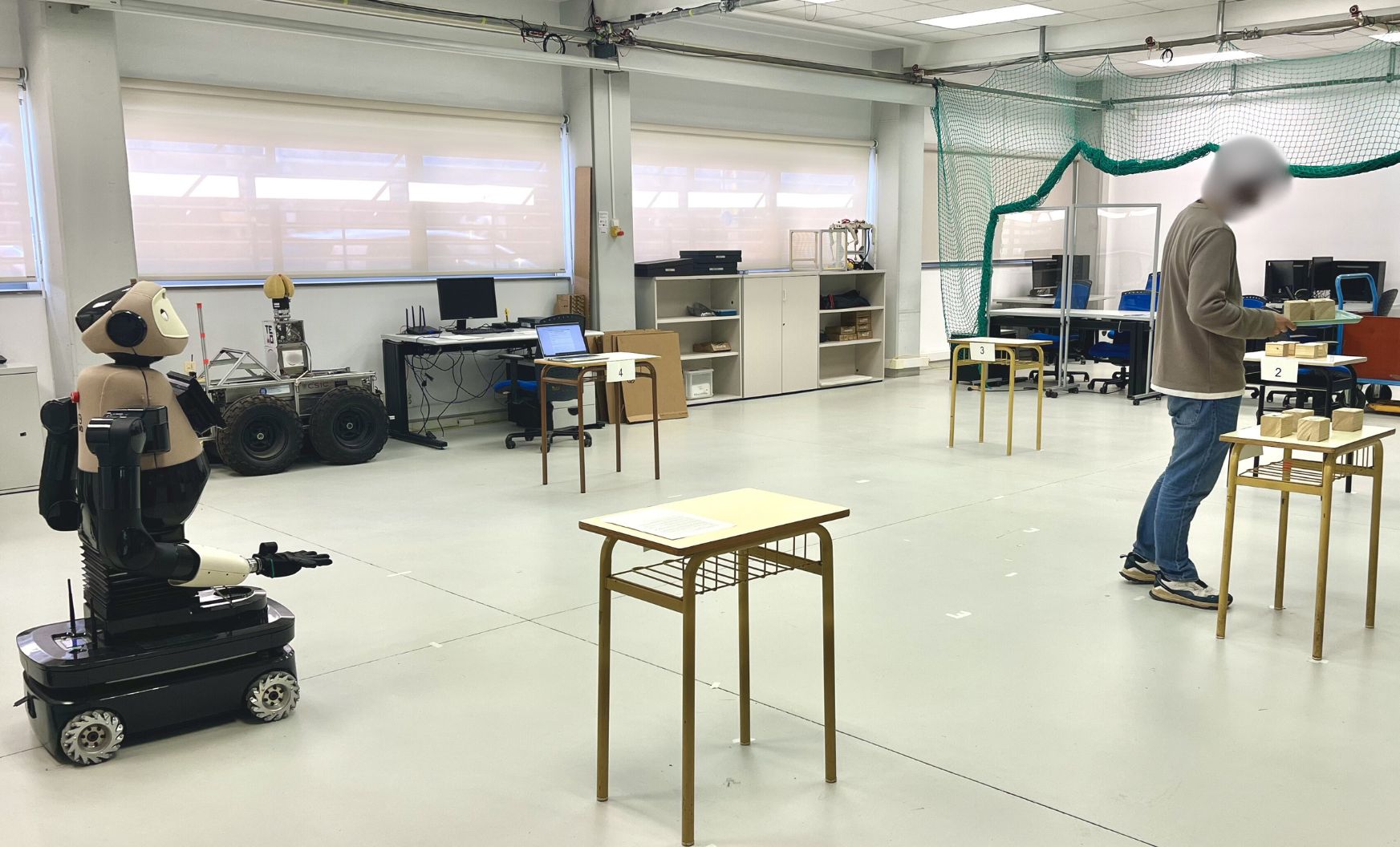}
    
    \vspace{0.02cm}
    
    \includegraphics[width=0.75\columnwidth]{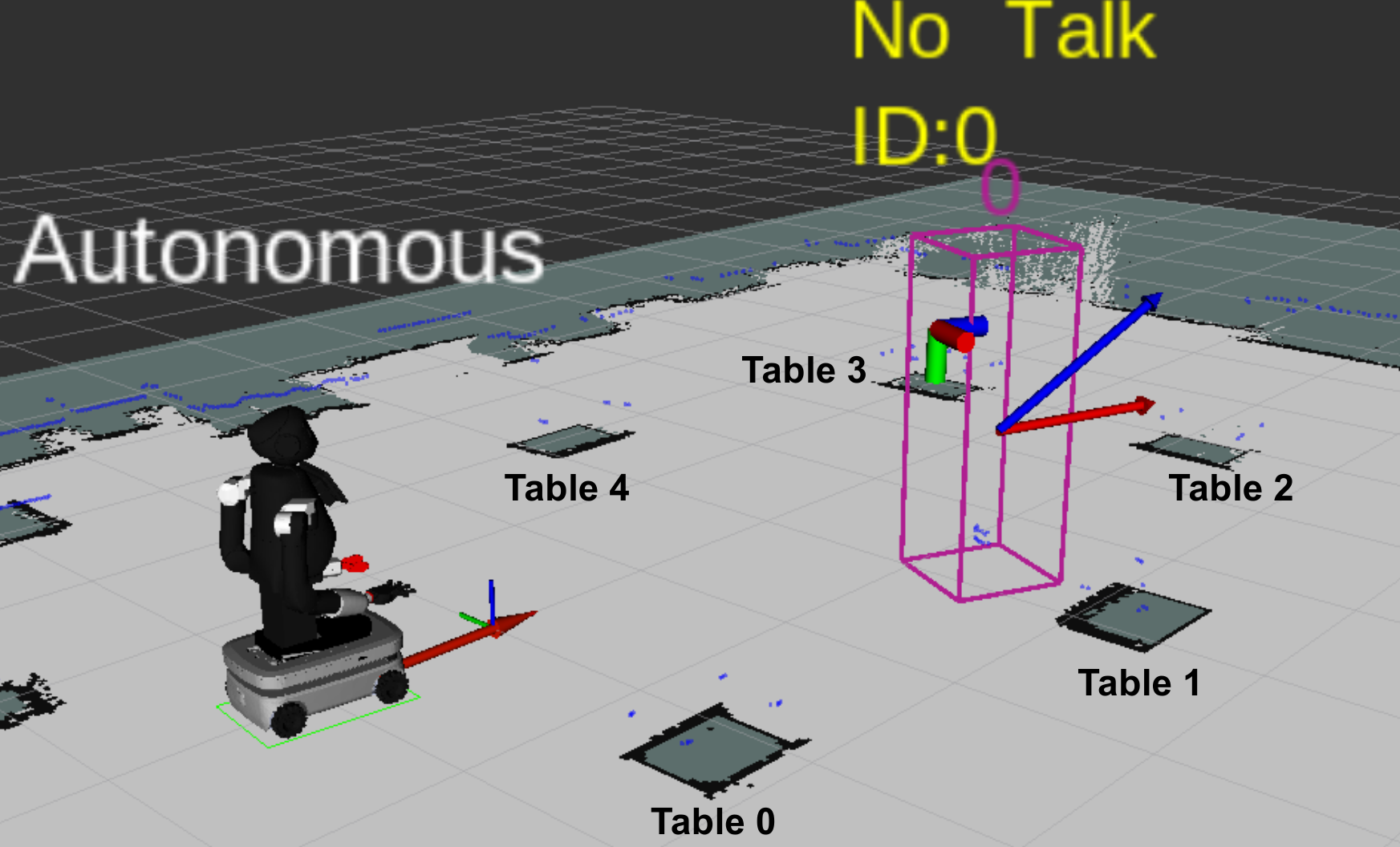}
    \caption{\textbf{Context-aware robot behavior.} \textit{Top}: Participant focused on the task, showing no intention to interact with the robot. \textit{Bottom}: RViz visualization of the robot’s perception.}
    \label{fig:no_talk}
    \vspace{-0.6cm}
\end{figure}

Despite increasing attention to engagement in HRI, few studies examine the negative effects of over-engagement.
In this work, we address the following research question: \textit{Does a robot that intervenes punctually, rather than continuously, provide a better overall user experience in task-oriented interactions within unfamiliar environments?}

To investigate this, we designed a task in which $32$ non-expert participants worked in a simulated hospital setting, preparing a patient-specific tray with a robot, as shown in~\fref{fig:no_talk}. The robot modulated its interventions using multimodal cues, including the user's position, head and body orientation, and verbal input.
Our contributions are twofold:
\begin{enumerate}
\item We design and implement two robot intervention policies: a robot-centric strategy that prioritizes user attention, and a context-aware strategy that balances human-robot engagement with task engagement.
\item We provide empirically grounded guidelines on how over-engagement relates to perceived intrusiveness and user experience in unfamiliar assistive environments.
\end{enumerate}

The paper is organized as follows. \sref{sec:rel_works} reviews related work, \sref{sec:method} describes the methodology, \sref{sec:exp_setup} presents the experimental setup, \sref{sec:results} provides the results, while \sref{sec:discussion} discusses them, and \sref{sec:conclusions} concludes the paper.

\section{Related Work} \label{sec:rel_works}


This section reviews engagement in HRI, covering assessment cues, robot adaptations, and human responses.

\subsection{Engagement Estimation}

Human expressiveness includes verbal and non-verbal signals of engagement, with the latter often providing more reliable but harder-to-interpret information. Audio and visual cues, such as gaze, head orientation, and body posture, are widely used indicators of engagement. 
These cues have been employed in rule-based approaches~\cite{foster2017automatically}, machine learning models~\cite{sidiropoulos2020measuring} and deep learning models~\cite{hanifi2024pipeline}.


Although gaze is a key cue, its reliability can decrease when users perform concurrent tasks. Models should therefore incorporate contextual information~\cite{castellano2012} and mechanisms to recover user's attention~\cite{sun2017sensing}.

\subsection{Adaptive Response to Engagement}


While many studies focus on engagement detection, few examine how this information can guide robot behavior to promote engagement~\cite {ravandi2025deep}. A common approach is to trigger re-engagement when low engagement is detected, sustaining participation and improving perceptions of the robot~\cite{cao2019hmm}. Even though proximity can increase attention~\cite{vazquez2014spatial}, it is not as relevant as robot performance~\cite{mead2016robots}.

Reinforcement learning has been used to encourage exploration of optimal robot behaviors~\cite{8404000}. In these systems, human evaluations of robot's social capabilities can serve as reward signals to maximize real-time engagement~\cite {del2020automatic}. However, adaptive behaviors have not always produced positive results, in some cases failing to increase engagement and even causing user annoyance~\cite{ziadeh2023feeling}.

Our study presents a novel comparison of robot intervention strategies to understand how humans perceive and prefer engagement during assistive tasks in unfamiliar environments.

\section{Methodology}\label{sec:method}

This section details the two HRI strategies compared in this study, developed using established re-engagement strategies~\cite{ravandi2025deep}. We first introduce the visual perception module used to track the user and infer engagement, and subsequently describe the control architectures that exploit this information to govern robot’s behavior.

\begin{figure}[t]
    \centering
    \includegraphics[width=0.72\columnwidth]{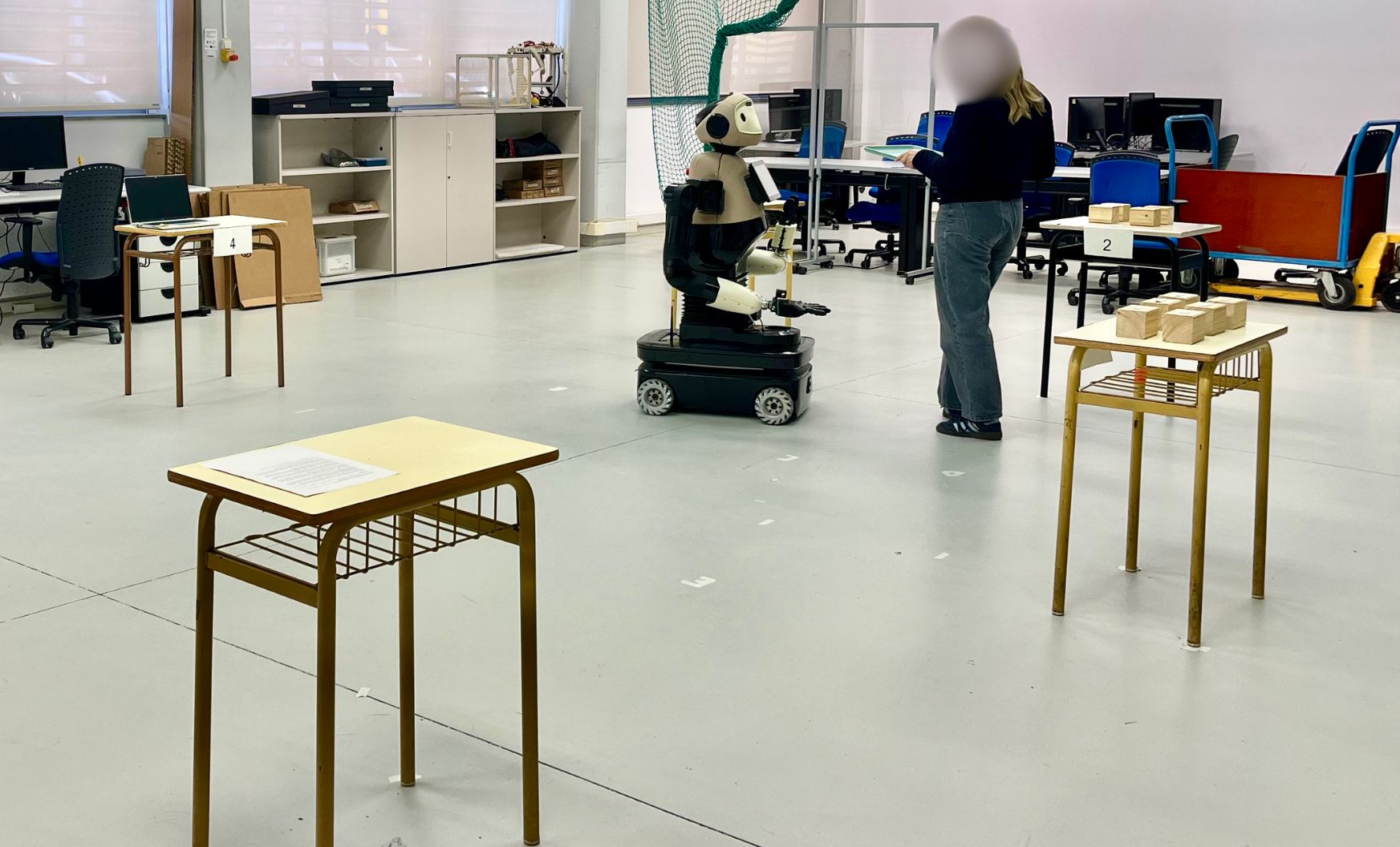}
    
    \vspace{0.03cm}
    
    \includegraphics[width=0.72\columnwidth]{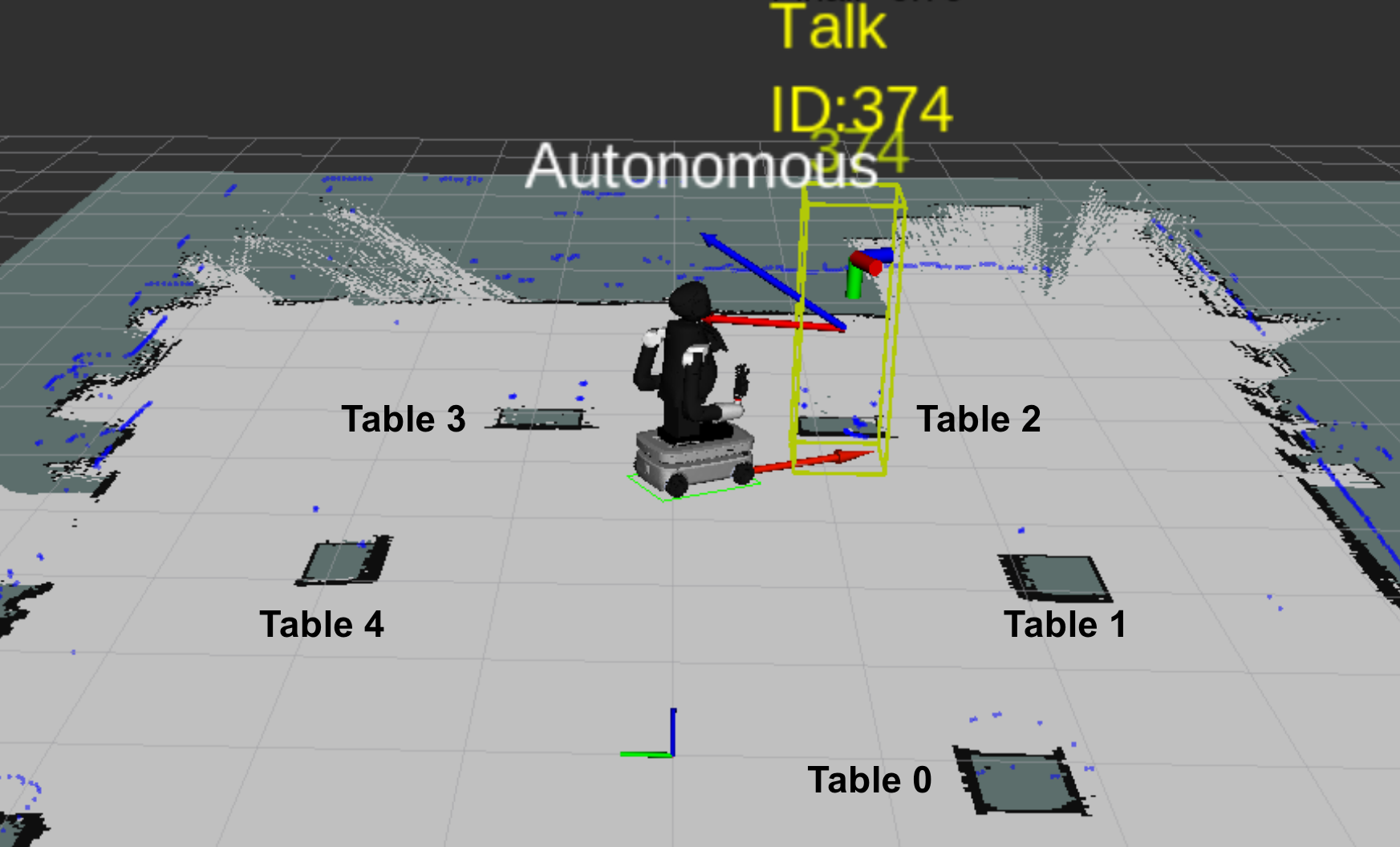}
    \caption{\textbf{Experimental setup.} \textit{Top}: Participant facing the robot, indicating an intention to interact. \textit{Bottom}: RViz visualization of the robot’s perception. The yellow bounding box marks the person, while the red and blue arrows indicate the torso and head directions, respectively.}
    \label{fig:talk}
    \vspace{-0.4cm}
\end{figure}

\begin{figure*}[t]
      \centering
      \includegraphics[width=\textwidth]{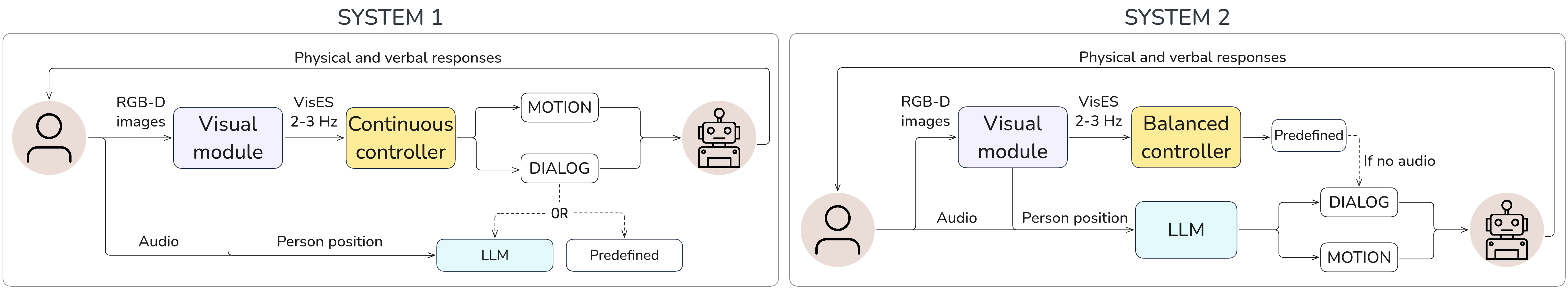}
      \caption{\textbf{Proposed systems.} \textit{Left}: continuous engagement system. \textit{Right}: balanced controller system. Both systems use visual input to track the person. In System 1, a continuous controller drives motion and dialogue, with the LLM invoked only under specific conditions. In System 2, the LLM controls responses and motion based on user input and visual context, with a balanced controller triggering occasional interventions when no audio input is detected.}
      \label{fig:systems}
   \end{figure*}
   
\subsection{Visual Module}

The visual module estimates the participant's position and extracts cues about their orientation and attention toward the robot. These signals serve as indicators of engagement, defined as the willingness to attend to and interact with a robot~\cite{bo2025hintbench, foster2017automatically}, and guide the robot’s intervention strategy.

Skeleton data are obtained from Mediapipe and regularized with a Kalman filter, yielding 3-D positions of key body landmarks. From this representation, participant's position in the environment is estimated, and three geometric indicators relative to the robot are extracted:

\begin{itemize}
\item Head orientation: estimated from facial landmarks to approximate the direction of the face normal, indicating whether the participant is visually attending to the robot.
\item Body orientation: computed from the vector connecting the left and right shoulders. The normal to this vector approximates the torso's forward direction and indicates whether the participant is oriented toward the robot.
\item Distance to the robot: computed as the Euclidean distance between the head position and the robot's frame.
\end{itemize}
This data is then merged and used to estimate an engagement score, which is employed to detect the \textit{NO TALK} or \textit{TALK} status, as shown in \fref{fig:no_talk} and \fref{fig:talk}, respectively.

\subsection{System 1: Continuous Engagement}
The first proposed system is presented in~\fref{fig:systems}. Following the idea that close distance can promote engagement~\cite{loos2024decoding}, the robot continuously follows the human in physical proximity.
When the user remains silent, any shift in orientation triggers re-engagement prompts that offer support. Likewise, changes in the user's position are interpreted as possible indications of requiring assistance, causing the robot to offer help whenever no conversation is currently in progress (e.g., "I will be here if you need help"). Each re-engagement prompt is delivered only once per state change; if the user does not reply, the robot returns to its default listening mode.

After a change in orientation or position, if the user speaks without any visual change, the LLM processes the first utterance. Subsequent dialogue is handled by the LLM only if the user is facing the robot; otherwise, the robot issues a predefined prompt to encourage attention (e.g., "I would appreciate it if you look at me when we are talking").
This design yields highly insistent behavior, with frequent prompts to couple user attention and dialogue progression.

\subsection{System 2: Balanced Engagement}

   
The second system, presented in~\fref{fig:systems}, implements a balanced engagement strategy, based on the idea that users performing a task may not continuously attend to the robot and should be engaged only when necessary. Unlike System 1, the robot does not follow the user at all times, and the LLM always processes verbal input regardless of the user's body orientation. The LLM not only generates verbal responses but can also trigger the robot to approach the user in response to explicit requests or contextual cues.

Orientation changes are interpreted as potential interaction signals. If the user shifts from not facing the robot to facing it without speaking, the robot issues a brief proactive prompt (e.g., "I can come if you want"). Similarly, key task locations may signal potential need for assistance; in these cases, situational prompts are issued when no verbal input is detected (e.g., "I think you have to answer some unknown data, do you want me to come and help you?").

Overall, this design produces adaptive, low-interruption behavior, assisting only when requested or contextually necessary, and avoiding proximity unless explicitly needed.

\section{Experimental Setup}\label{sec:exp_setup}

In this section, we describe the study design, including the technical implementation of the robot systems, the task setup, participant details, and the measures and questionnaires used.

\subsection{Technical Interaction Design}

The experiments were conducted using IVO~\cite{laplaza}, a bimanual mobile robot presented in~\fref{robot_screen}. System 1 and System 2 continuously monitored visual data, including the user's position, body, and head orientation, while also processing verbal input \cite{hriscu2025human}. The robot aligned its head with the user's pose throughout the interaction. Listening was the default state, with audio input transcribed using the Speech-to-Text (STT) English Vosk model.
Both systems employed the LLM Qwen-2.5:7B-Instruct to answer user queries, while receiving up-to-date previous dialog and positional information from the visual module to enable context-aware responses. Depending on the system-specific rules, the robot either initiated interactions or triggered re-engagement behaviors. Robot-initiated utterances without prior user input were randomly selected from a predefined set specific to each situation and system. All spoken responses were synthesized using Piper, a Text-to-Speech (TTS) model.

\subsection{Experiment Design}

\begin{figure}[t]
 \centering
  \subfloat[]{
\hspace{-0.3cm}
   \label{robot_screen}
    \includegraphics[width=0.47\columnwidth]{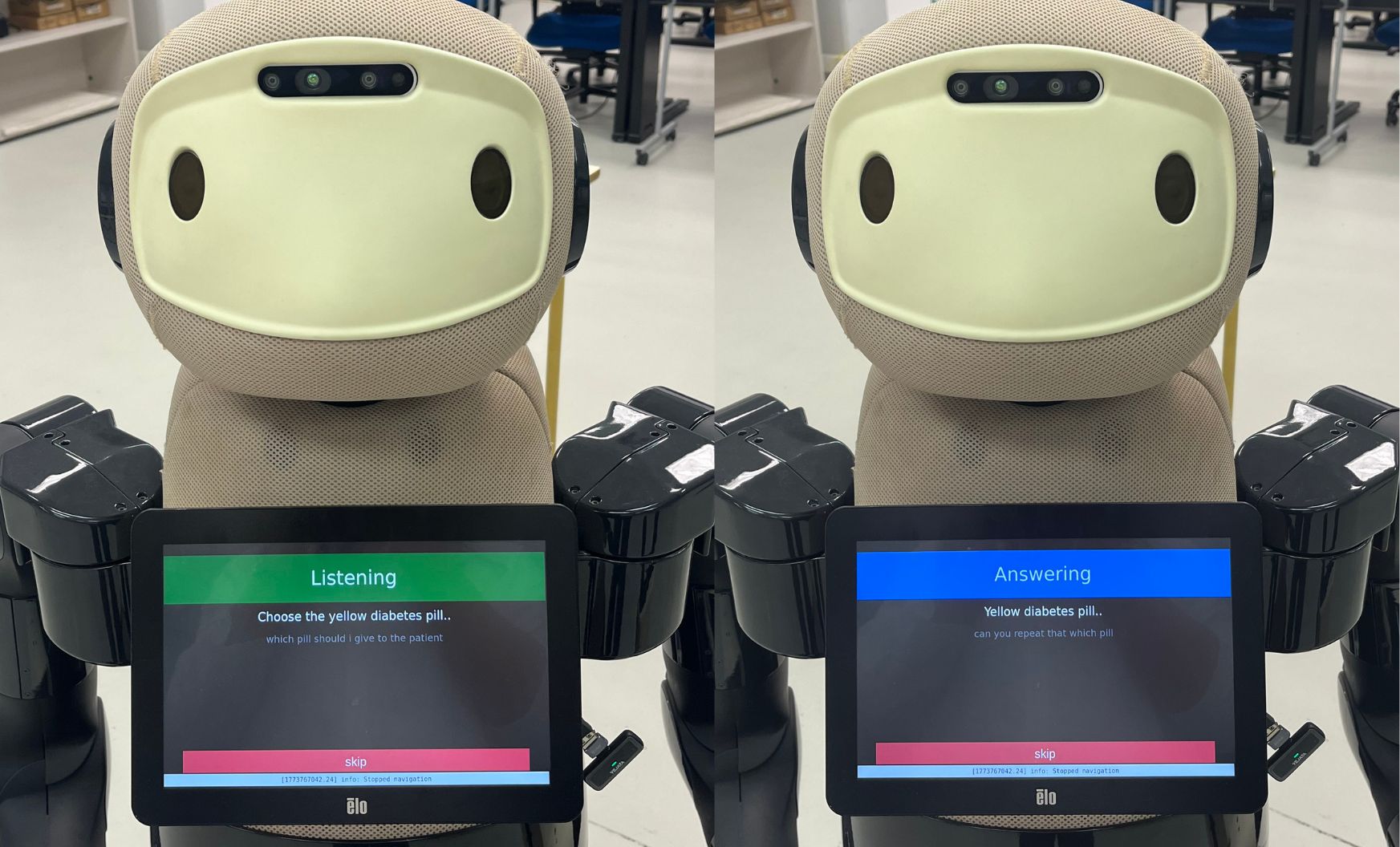}}
  \subfloat[]{
    \hspace{-0.3cm}
   \label{food}
    \includegraphics[width=0.47\columnwidth]{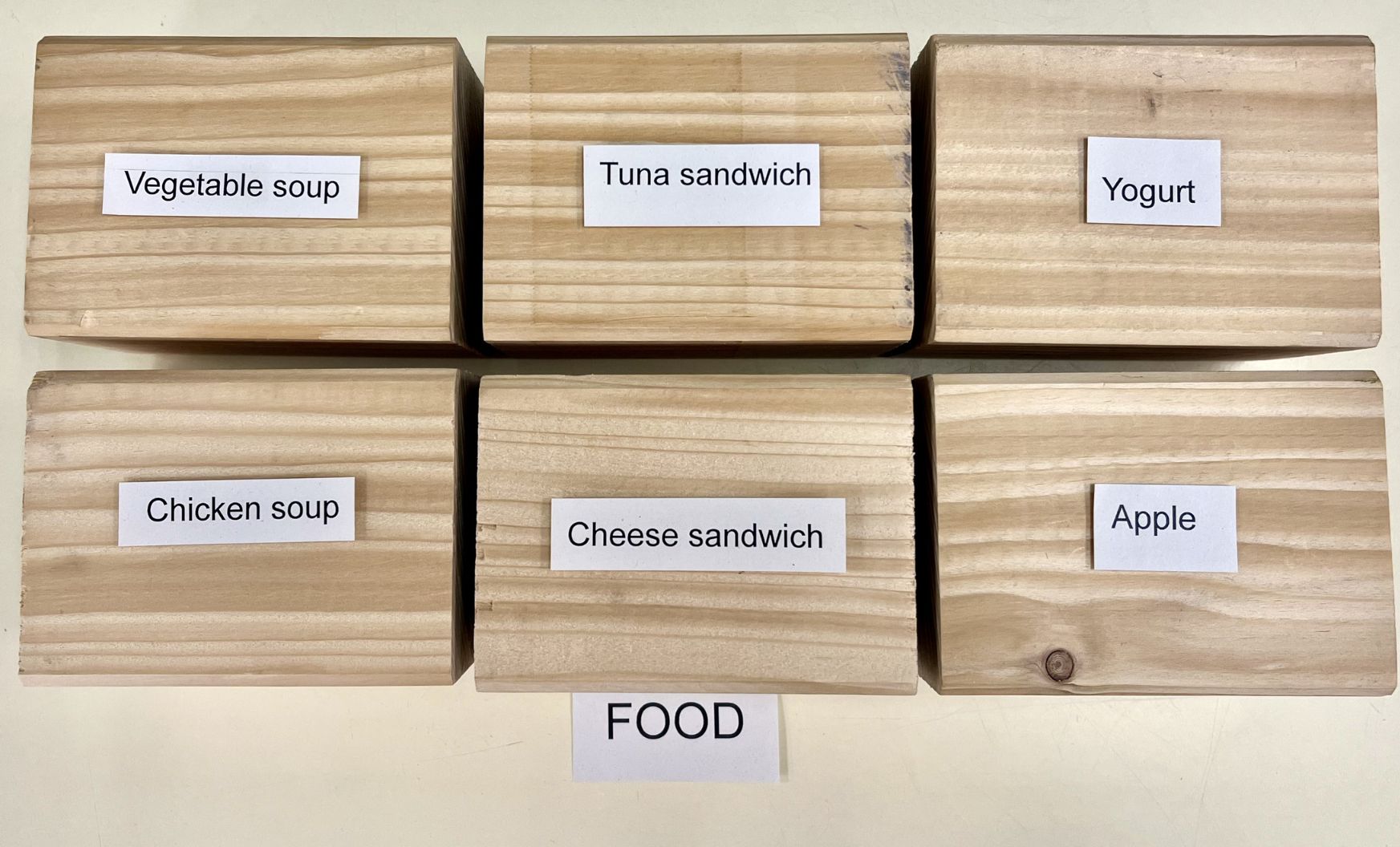}}\\ \vspace{0.1cm}
  \subfloat[]{
  \hspace{-0.3cm}
   \label{fluids}
    \includegraphics[width=0.47\columnwidth]{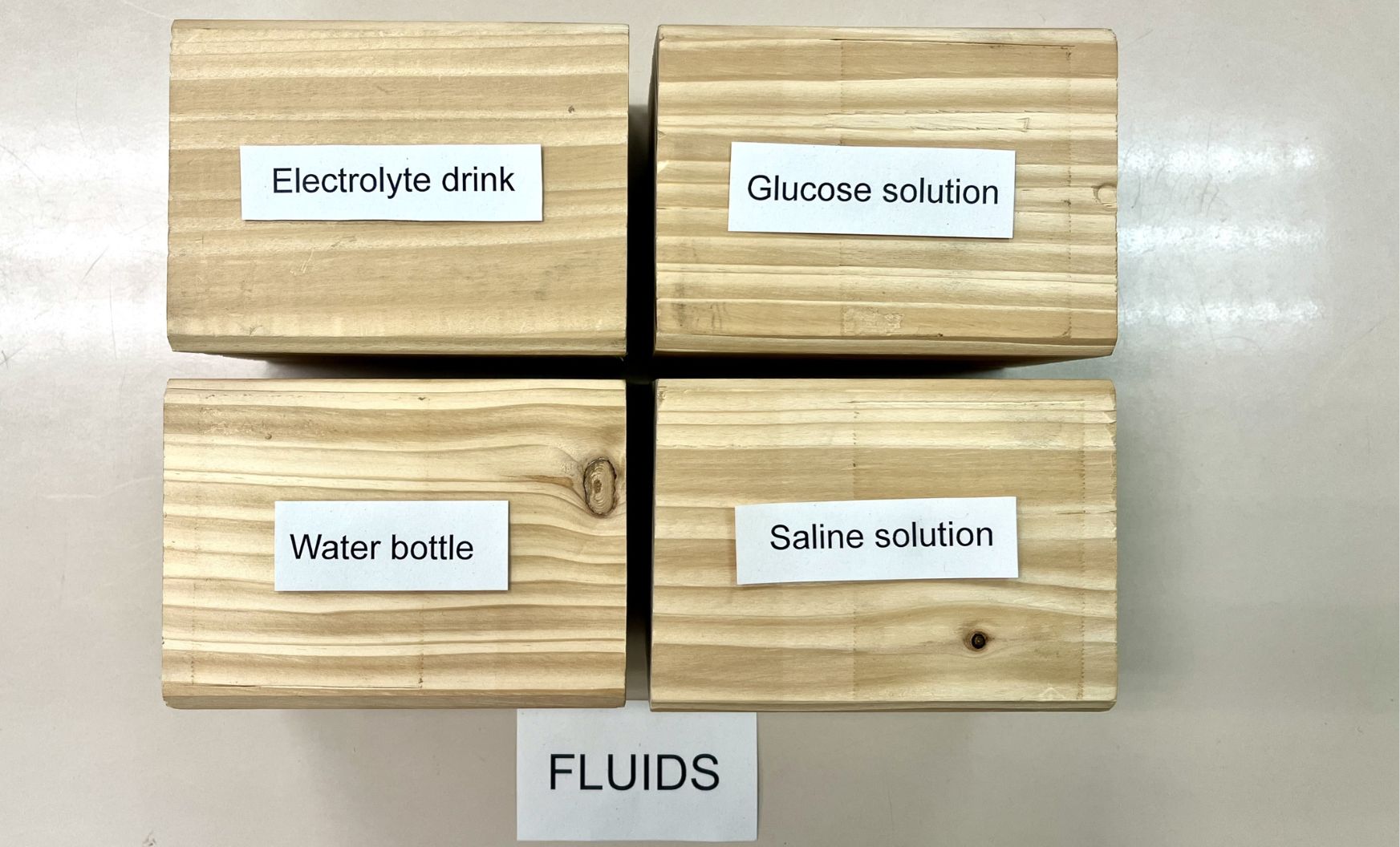}}
    \subfloat[]{
 \hspace{-0.3cm}
   \label{pills}
    \includegraphics[width=0.47\columnwidth]{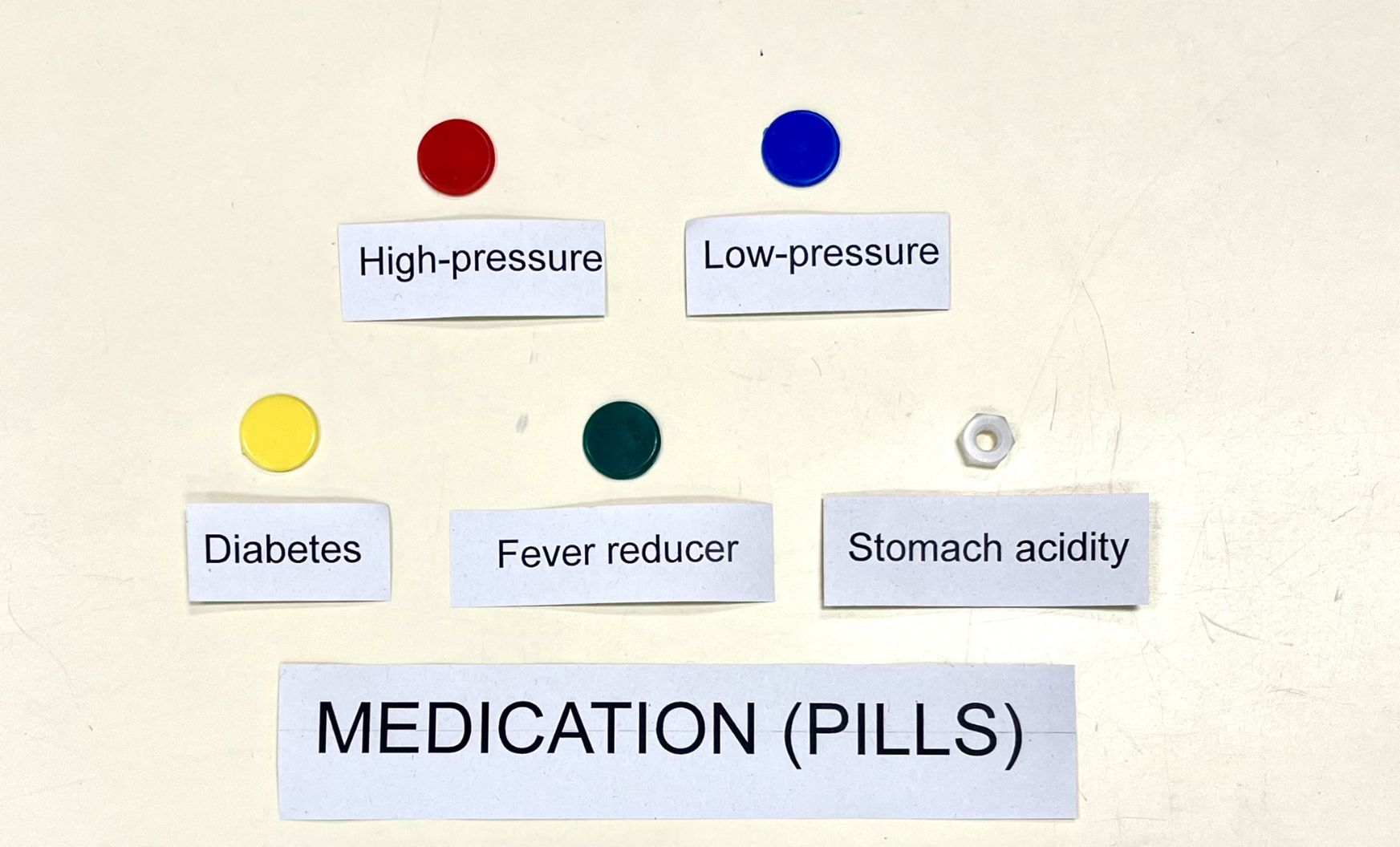}}
 \caption{\textbf{Robot and task items.} \textit{(a)} Robot screen showing listening (green) and responding (blue) states, along with dialog transcripts; a red skip button interrupts responses and resumes listening. \textit{(b)} Food, \textit{(c)} Fluids, \textit{(d)} Medication pills.}
 \label{fig:real_scenario}
 \vspace{-0.4cm}
\end{figure}

Participants performed a task in a simulated hospital warehouse environment, assuming the role of a nurse preparing the items shown in~\fref {fig:real_scenario} for a patient. As shown in~\fref{fig:no_talk} and~\fref{fig:talk}, the environment contained five tables: an initial table with instructions and patient data; three tables with different item categories; and a final table for placing the selected items and completing a registration form. Each participant received a patient information sheet containing physiological values, health conditions, and dietary restrictions, along with instructions specifying the number of items to collect from each category, but not which specific items to choose. The task was designed so that some decisions required interpreting medical information (e.g., blood pressure values or dietary restrictions), creating natural uncertainty and opportunities for robot assistance. Importantly, even if some items were not recommended for the specific patient, there was no single correct combination of items, as the study did not aim to evaluate accuracy.
To ensure that each participant interacted with the robot at least once, the final step was to complete a confirmation form at the last table. This form requested the patient's identification code, available only by asking the robot.

At the beginning of the task, the robot explained the procedure to the participant, pointed out the patient sheet, and indicated that it was the sole available source of help. Participants could move freely among the tables and select items according to the instructions. They carried out the task twice, with the systems and the two patients randomized. 


\subsection{Participants}


Before the experiment, participants completed a questionnaire collecting demographic information. A total of $32$ volunteers participated in the study ($19$ male, $13$ female), ranging in age from $19$ to $65$ years ($M = 31.22$, $SD = 11.18$). Participants were not affiliated with hospitals, so they were in an unfamiliar setting. After each interaction, they answered questions about their experience and their preferred system.
All participants signed a consent form approved by our institution's Ethics Committee (ID: 2026-054).

\subsection{Measures}

After each interaction, participants evaluated their experience through several measures: emotional response using the Self-Assessment Manikin (SAM) \cite{bradley1994measuring}, perceived intelligence via the Godspeed questionnaire \cite{bartneck2009measurement}, engagement using the User Engagement Scale (UES) \cite{o2018practical}, and perceived intrusiveness on a 5-point Likert scale using study-specific items:

\begin{enumerate}
    \item The robot interrupted me unnecessarily.
    \item The robot invaded my personal space.
    \item The robot intervened at inappropriate moments.
    \item The robot respected my autonomy (reverse-coded).
\end{enumerate}

After completing both conditions, participants indicated their preferred interaction (based on the robot's movement and speech) and optionally provided written justifications and additional feedback. Concerning our research question, we present three main hypotheses:

\begin{itemize}
    \item H1 - User Preference: Participants prefer the balanced engagement over the continuous engagement system.
    \item H2 - Perceived Intelligence: Both systems are perceived as equally intelligent, providing comparable assistance.
    \item H3 - Perceived Intrusiveness: The balanced system is perceived as less intrusive than the continuous one.
\end{itemize}

We also collected interaction logs, including conversations, durations, user and robot message counts, and participant position changes (table switches).

\section{Results}\label{sec:results}

This section presents the experimental findings, divided into user preferences, experience, and interaction metrics.

\subsection{User Preferences Data}

Participants did not have a dominant preference for one system over the other, as $14$ preferred System 1, $15$ preferred System 2, and $3$ reported no clear preference.

Considering order of exposure, among the $16$ participants who first interacted with System 1, $2$ reported no preference, $6$ preferred System 1, and $8$ preferred System 2. Similarly, among the $16$ participants who first interacted with System 2, $1$ reported no preference, $8$ preferred System 1, and $7$ preferred System 2. These values suggest a negligible and non-systematic effect of interaction order on preference.

To test H1, we conducted a one-sided binomial test ($p = 0.5$) excluding neutral responses to assess whether System 2 was preferred over System 1, given the expectation that the latter would be perceived less favorably. No significant difference in preference was found between systems.

\subsection{User Experience for Each System}

To assess how system differences affected user perceptions, we analyzed emotional responses using the SAM questionnaire, engagement using the UES, and perceived intelligence and intrusiveness of each system.

\paragraph{Emotional Response}

For valence, System 2 elicited slightly more positive ratings ($M = 6.34$, $SD = 2.03$) compared to System 1 ($M = 6.06$, $SD = 1.70$). Arousal was slightly higher for System 1 ($M = 4.72$, $SD = 2.43$) than System 2 ($M = 4.16$, $SD = 2.40$), indicating greater perceived activation and excitement when interacting with System 1. Finally, dominance scores were comparable across conditions, with a small increase for System 2 ($M = 6.19$, $SD = 1.97$) relative to System 1 ($M = 5.84$, $SD = 1.89$).
We also examined the effect of interaction order, which revealed no meaningful differences across conditions, with variations below $0.25$ on all metrics.

\paragraph{Engagement}

Results showed very similar engagement levels across conditions, with System 1 ($M = 3.64$, $SD = 0.60$) slightly higher than System 2 ($M = 3.57$, $SD = 0.69$), but the difference was not meaningful.
No order of exposure effects were observed on UES.
While System 1 showed a slight decrease in UES from first to second interaction, System 2 showed a slight increase, suggesting that engagement fluctuations across sessions were inconsistent and likely due to noise rather than a structured effect.

\paragraph{Perceived Intelligence}

Scores were similar across conditions, with System 1 ($M = 3.71$, $SD = 0.76$) and System 2 ($M = 3.64$, $SD = 0.78$). We also found consistent values across interaction order (first session: $M=3.78$, $SD=0.62$; second session: $M=3.56$, $SD=0.87$).
A Shapiro-Wilk test confirmed normality of the paired differences ($p = 0.311$), and a paired-samples t-test was conducted to evaluate differences in perceived intelligence between systems. The results showed no significant effect ($t(31) = 0.41$, $p = 0.68$), indicating that the null hypothesis of H2 could not be rejected.

\begin{table}
\centering
\caption{\textbf{Experiment metrics}. Average values measured.}
\begin{tabular}{lcc}
\toprule
\textbf{System} & \textbf{S1} & \textbf{S2}
\\
\midrule
\textbf{Duration (s)} & $303.2 \pm 91.5$ & $342.3 \pm 113.3$\\
\textbf{Position changes} & $7.6 \pm 3.1$ & $11.3 \pm 5.5$\\
\textbf{User Inputs} & $18.3 \pm 9.9$ & $21.8 \pm 10.4$ \\
\textbf{System Outputs} & $28.1 \pm 10.5$  & $25.8 \pm 11.5$\\
\bottomrule
\end{tabular}
\label{tab:objective_metrics}
\vspace{-0.5cm}
\end{table}

\paragraph{Perceived Intrusiveness}

Concerning H3, System 1 was rated as more intrusive ($M = 2.88$, $SD = 1.08$) than System 2 ($M = 1.91$, $SD = 0.79$). No clear order effects were observed (first session: $M = 2.24$, $SD = 0.92$; second session: $M = 2.55$, $SD = 1.19$), suggesting that perceptions of intrusiveness were primarily driven by system behavior.
Normality of the paired differences was confirmed using the Shapiro-Wilk test ($p = 0.650$), and a paired-samples t-test revealed a statistically significant difference in perceived intrusiveness between the systems ($t(31) = 4.83$, $p = 3.49\times10^{-5}$). This indicates that the balanced engagement system was perceived as significantly less intrusive.

\subsection{Interaction Metrics}\label{sec:objective}

The logs of the experiments are reported in  \tref{tab:objective_metrics}. System 1 produced shorter interactions with moderate user movement. In contrast, System 2 exhibited longer interactions, greater movement, more user inputs, and fewer robot outputs.

\begin{figure}[t]
    \centering
    \subfloat[\label{fig:duration_corr}]{
        \includegraphics[width=0.91\columnwidth]{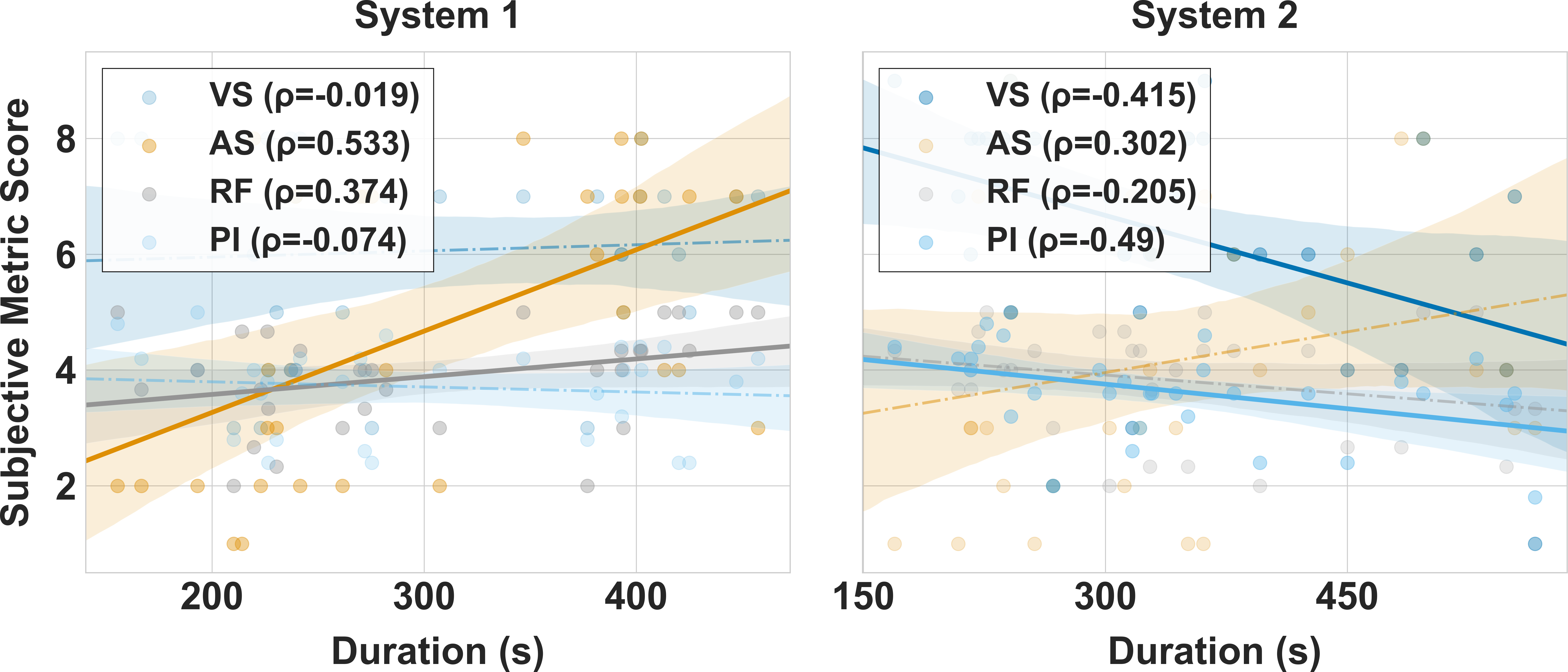}
    } \\
    \subfloat[\label{fig:input_corr}]{
        \includegraphics[width=0.91\columnwidth]{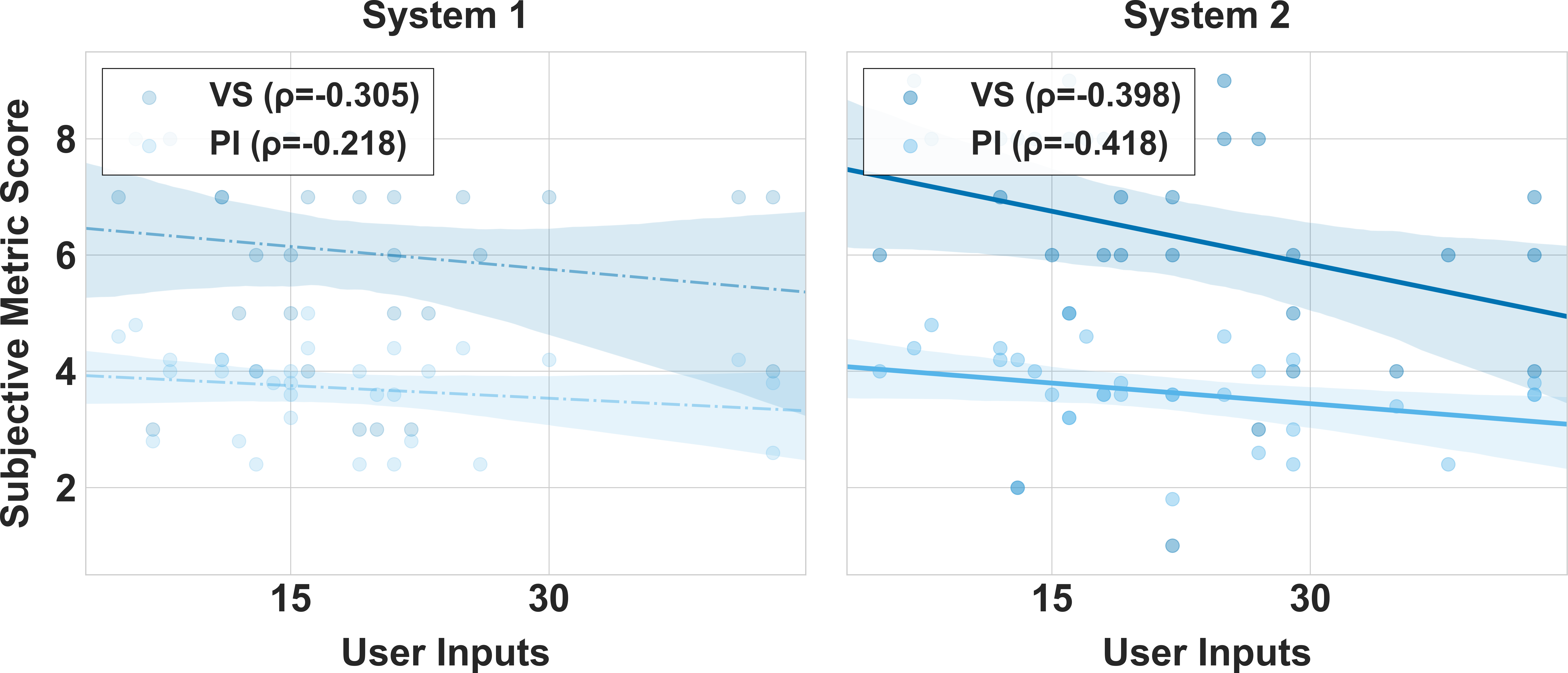}
    }\\
    \subfloat[\label{fig:output_corr}]{
        \includegraphics[width=0.91\columnwidth]{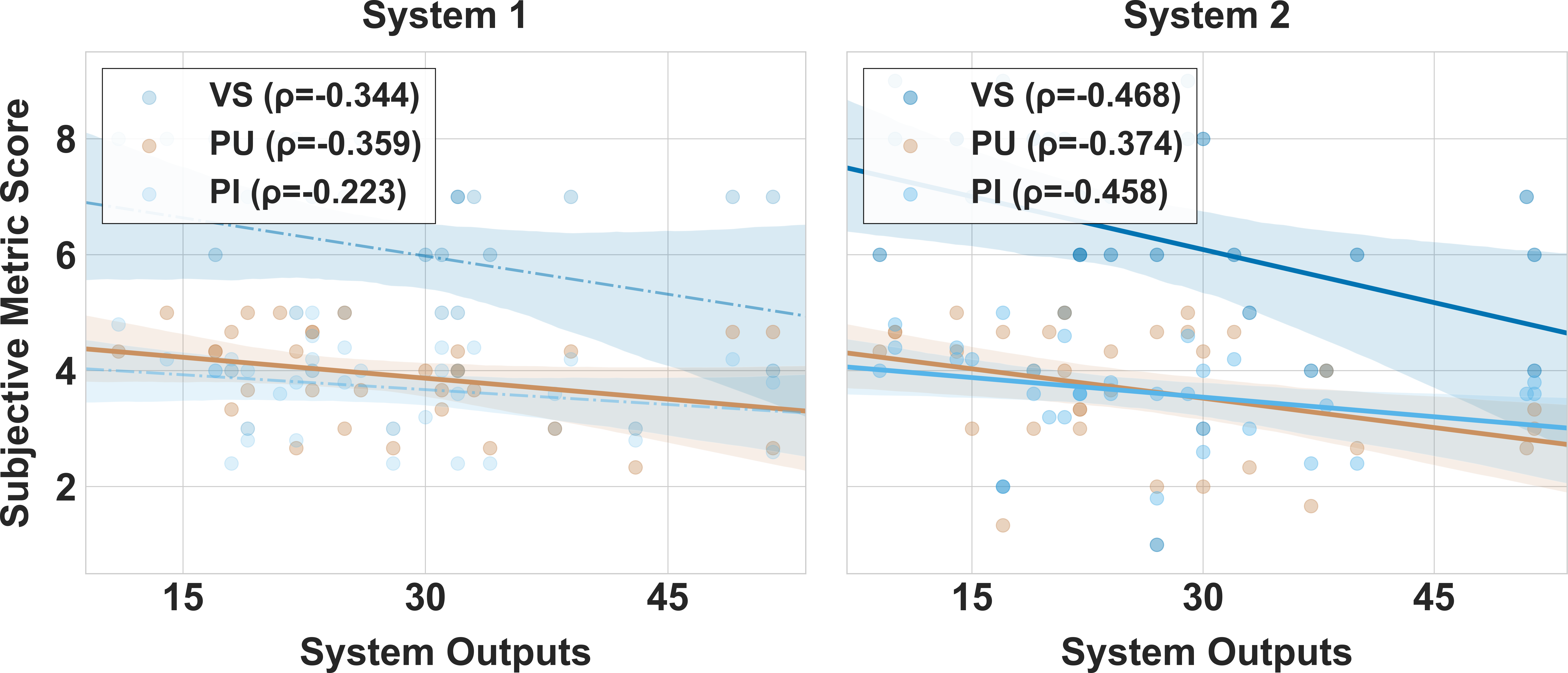}
    }

    \caption{\textbf{Correlations between objective and subjective metrics.} Scatter plots showcasing correlation for both systems, grouped by objective metric. Dashed lines and solid lines indicate non-significant and significant correlations ($p < 0.05$), respectively. Legend with statistics' values: Valence Score (VS), Arousal Score (AS), Reward Factor (RF), Perceived Intelligence (PI), and Perceived Usability (PU).}
    \label{fig:correlation_obj_subj}
    \vspace{-0.4cm}
\end{figure}

We conducted a Spearman correlation analysis on the subjective questionnaire metrics, as normality was not met, including interaction number to account for potential order effects. Significant results are reported in \fref{fig:correlation_obj_subj}. For System 2, longer durations and more outputs correlated negatively with valence and perceived intelligence, suggesting that greater interaction intensity may reduce subjective enjoyment or perceived robot competence. System 1 shows fewer significant correlations, but increased duration is positively related to arousal and reward factor, and higher outputs reduce perceived usability, consistent with the system's more intrusive, intervention-heavy behavior. No significant order effects were found, indicating successful randomization.

\section{Discussion}\label{sec:discussion}

In this section, we discuss the most relevant findings and relate them to participants' feedback.

\subsection{Balancing Support vs. Intrusiveness}

We noticed that increased robot intervention does not linearly improve user experience, as neither system was consistently preferred. Instead, engagement preferences were shaped by perceived robot performance and how support was framed, consistent with~\cite{mead2016robots}.
Participants who preferred System 2 emphasized autonomy and reduced pressure, as frequent robot interventions felt overbearing and undermined their sense of competence. This suggests that continuous assistance may be overly directive rather than supportive. 

In contrast, participants who preferred System 1 acknowledged its greater intrusiveness but described it as offering stronger companionship, particularly in unfamiliar environments. Intrusiveness may be perceived as a supportive presence rather than a negative trait, depending on the context.

Given that both systems differed only in intervention strategy, the comparable perceived intelligence suggests that users primarily evaluate intelligence based on task effectiveness and response quality rather than on interaction strategy.

\subsection{Engagement in Task-Oriented Interactions}

Although overall engagement scores were similar across conditions, our findings highlight important qualitative differences in interaction dynamics. System 2 fostered a more user-driven interaction style, characterized by longer task durations, increased user input, and greater physical exploration. While this encouraged active participation, participants reported feeling less supported and more isolated, suggesting that higher user effort can undermine user experience.

Moreover, increased interaction load was negatively associated with valence and perceived intelligence, indicating that sustained effort may diminish user experience. In contrast, System 1 promoted shorter, more reactive exchanges, associated with higher excitement but also with signs of potential over-intervention, as higher system output frequency correlated with reduced perceived usability.
Overall, engagement should be understood not simply as interaction volume, but as an optimal balance between user effort and system initiative.

\subsection{Physical Presence and Social Behavior in HRI}

Robot proximity significantly shaped interaction since participants felt more accompanied when the robot was closer, which encouraged more structured, direct queries. These clearer inputs improved downstream language model performance, particularly in task-oriented settings where ambiguity can degrade response quality~\cite{10825265}. Most participants found the robot’s proximity unnecessary and invasive, with some perceiving unsolicited approaches as intrusive. Incorporating permission-aware movement strategies is therefore essential.

Participants also naturally applied human social norms, frequently orienting themselves toward the robot and moving closer rather than interacting at a distance or asking the robot to move. This suggests an intuitive adaptation to face-to-face conversational conventions even when unnecessary.
Finally, proximity and embodied presence appeared to influence affective responses, with some participants describing System 2 as more relaxed and reporting slightly lower arousal, highlighting the importance of adapting robot behavior to emotional responses. Proximity also improved interface visibility and facilitated interaction.

\section{Conclusions and Future Work}\label{sec:conclusions}
In this work, we experimented with 32 volunteers to investigate how different robot intervention strategies influence user engagement, perception, and behavior in task-oriented Human-Robot Interaction (HRI). We compared a continuous, robot-centric engagement strategy with a context-aware approach designed to balance user attention, autonomy, and task demands.
Our findings suggest that robot intervention in unfamiliar tasks requires a balance between user autonomy and social presence. Intrusiveness can be verbal and physical and can be associated with perceived support. Engagement quality is not equivalent to interaction quantity, and it is shaped by how system initiative and user effort are distributed, as well as by physical embodiment and proximity.

Future work will focus on integrating these findings into adaptive frameworks that personalize robot behavior based on user characteristics and contextual factors. This development will contribute toward more effective, user-aware robotic systems for real-world applications.





\bibliographystyle{IEEEtran}
\bibliography{references}

\end{document}